%% file: main.tex
\documentclass[letterpaper, 10 pt, conference]{ieeeconf}  
\usepackage{textgreek}

\IEEEoverridecommandlockouts                              

\usepackage{tcolorbox}
\usepackage[utf8]{inputenc}
\usepackage[T1]{fontenc}
\usepackage{bm}
\usepackage{graphicx}
\usepackage{color,soul}
\usepackage{dblfloatfix}
\usepackage{multirow}
\usepackage{multicol}
\usepackage{hyperref}
\usepackage{siunitx}
\usepackage{hyperref}
\usepackage{xcolor}
\usepackage{graphicx}
\usepackage{svg}
\usepackage{amsmath}
\usepackage{amsfonts}
\usepackage{amssymb}
\usepackage{array}
\usepackage{url}
\usepackage{epstopdf}
\usepackage{float}
\usepackage{subfigure}
\usepackage{threeparttable}
\usepackage{bm}
\usepackage{caption}
\usepackage{algorithm,algorithmic}
\usepackage{booktabs}
\usepackage{tabularx}
\usepackage[utf8]{inputenc} 
\usepackage[T1]{fontenc}    
\usepackage{hyperref}       
\usepackage{url}            
\usepackage{booktabs}       
\usepackage{amsfonts}       
\usepackage{nicefrac}       
\usepackage{microtype}      
\usepackage{xcolor}         
\usepackage{graphicx}
\usepackage{multirow}
\usepackage{graphicx}
\usepackage{bbding}
\usepackage{colortbl}
\usepackage{CJKutf8}
\usepackage{cite}
\usepackage{subcaption} 
\usepackage[utf8]{inputenc} 
\DeclareUnicodeCharacter{2212}{\ensuremath{-}}

\begin{document}

\definecolor{darkgreen}{rgb}{0.0, 0.5, 0.0}
\definecolor{darkred}{rgb}{0.5, 0.0, 0.0}

\newcommand{\etal}{\textit{et al}.}

\title{\LARGE \bf CollabVLA: Self-Reflective Vision–Language–Action Model\\ Dreaming Together with Human}

\author{Nan Sun$^{1,*}$,   
 Yongchang Li$^{1,*}$, Chenxu Wang$^{1}$, Huiying Li$^{1}$ and Huaping Liu$^{1,\dagger}$%
\thanks{This work was supported by National Natural Science Fund under Grants 62025304 and 62273054. $^{*}$ denotes the equal contribution.}
\thanks{$^{1}$The author is with the Department of Computer Science and Technology, Tsinghua University, Beijing, 100084, China.}%
\thanks{$^\dagger$Corresponding Authors. {hpliu@tsinghua.edu.cn}}%
}

\maketitle
\thispagestyle{empty}
\pagestyle{empty}


\begin{abstract}
In this work, we present CollabVLA, a self-reflective vision–language–action framework that transforms a standard visuomotor policy into a collaborative assistant. CollabVLA tackles key limitations of prior VLAs, including domain overfitting, non-interpretable reasoning, and the high latency of auxiliary generative models, by integrating VLM-based reflective reasoning with diffusion-based action generation under a mixture-of-experts design. Through a two-stage training recipe of action grounding and reflection tuning, it supports explicit self-reflection and proactively solicits human guidance when confronted with uncertainty or repeated failure. It cuts normalized \emph{Time} by $\sim\!2\times$ and \emph{Dream} counts by $\sim\!4\times$ vs.\ generative agents, achieving higher success rates, improved interpretability, and balanced low latency compared with existing methods. This work takes a pioneering step toward shifting VLAs from opaque controllers to genuinely assistive agents capable of reasoning, acting, and collaborating with humans.

\end{abstract}

\section{Introduction}

Large-scale vision--language models (VLMs) excel at open-world perception and instruction-following~\cite{alayrac2022flamingovisuallanguagemodel,zhai2023sigmoidlosslanguageimage}, motivating vision-language-action (VLA) policies that fine-tune on robot data via autoregressive next-token prediction~\cite{brohan2023rt1roboticstransformerrealworld,kim2024openvlaopensourcevisionlanguageactionmodel,li2024visionlanguagefoundationmodelseffective}. Yet this often degrades multimodal grounding due to domain overfitting, and scarce robot data prevents scaling generalization as in LLMs. To address this, some methods co-train VLAs on paired image--text corpora and learn latent action from internet videos~\cite{brohan2023rt2visionlanguageactionmodelstransfer,liang2025clamcontinuouslatentaction,chen2025villaxenhancinglatentaction}. However, co-training risks task interference~\cite{zhou2025chatvlaunifiedmultimodalunderstanding}, while latent actions, though compact, remain non-interpretable and add training complexity~\cite{liang2025clamcontinuouslatentaction}.

Inspired by chain-of-thought reasoning in LLMs~\cite{wei2023chainofthoughtpromptingelicitsreasoning}, we naturally ask whether VLAs can also “think step-by-step” by making intermediate reasoning explicit. Indeed, recent work has explored this direction through textual planning, visual subgoal, and auxiliary prediction~\cite{zawalski2025roboticcontrolembodiedchainofthought,zhao2025cotvlavisualchainofthoughtreasoning,zhou2024robodreamerlearningcompositionalworld,li2024grmgleveragingpartiallyannotated}, showing improved transparency and generalization by aligning high-level reasoning with low-level policy.

Yet these methods often fall short. For compact open-source VLMs or world models, explicit subgoal generation, particularly photorealistic egocentric images, usually generalizes only to seen layouts, while adding latency. Besides, their explicit reasoning remains largely a surface-level narration of the current situation, functioning as an internal monologue with limited guidance value. It lacks the deeper reflective understanding necessary for real-time failure recognition and effective interaction, as illustrated by the evolution from ReAct~\cite{yao2023reactsynergizingreasoningacting} to Reflexion~\cite{shinn2023reflexionlanguageagentsverbal} in LLM agents.

\begin{figure}[t]
    \centering
    \includegraphics[width=1\linewidth]{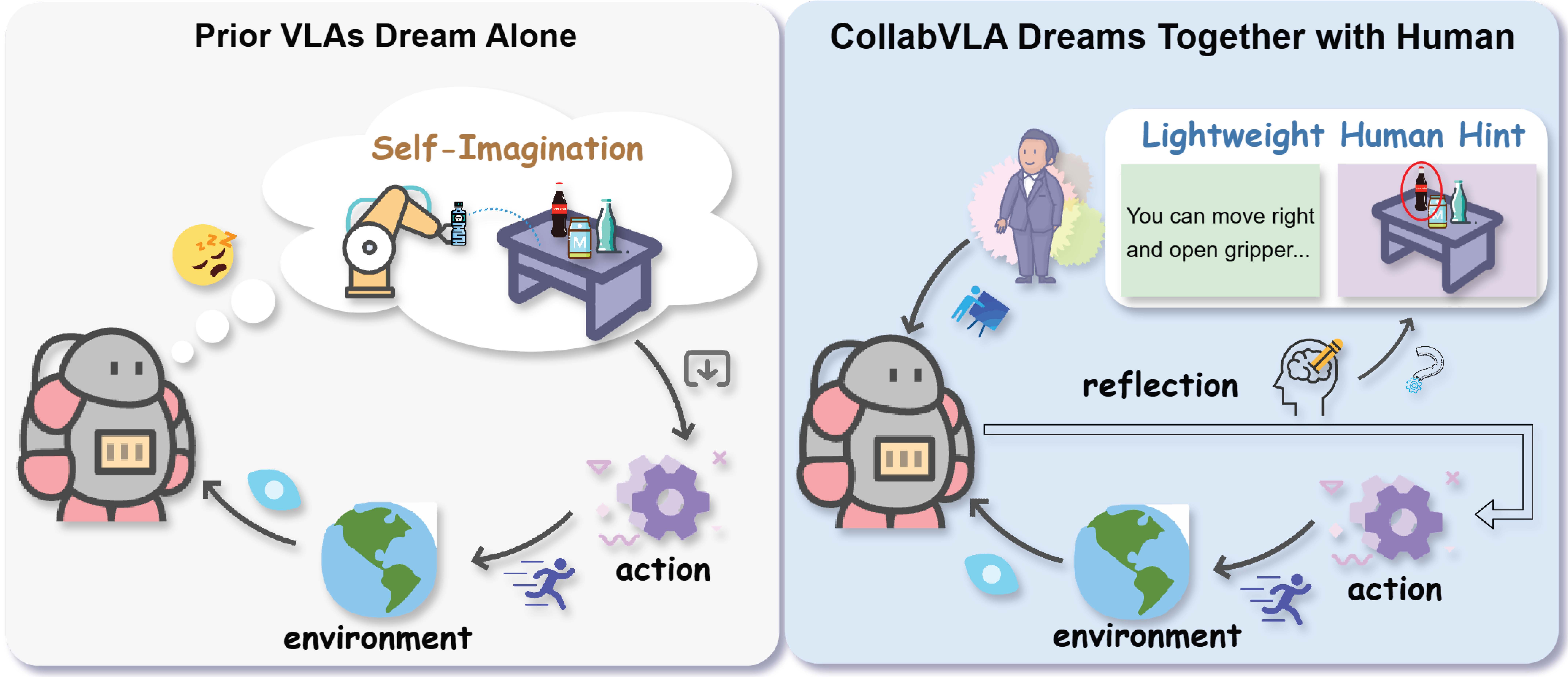}
    \caption{CollabVLA extends beyond self-imagination by integrating human guidance with action generation, transforming a closed-loop visuomotor policy into a collaborative agent.}
    \label{fig:introduction}
    \vspace{-5mm}
\end{figure}

In this work, we present \textbf{CollabVLA}, a collaborative VLA capable of self-reflection and can proactively seek human guidance, rather than relying solely on inefficient and imperfect self-imagination (see Fig.~\ref{fig:introduction}). It integrates autoregressive VLM-based language generation with diffusion-based action generation under a mixture-of-experts (MoE) adaptation. CollabVLA follows a two-stage training recipe: (1) \textit{Action Grounding}, where a VLM-driven action policy is trained on latent action representations conditioned on multimodal goals to master acting. To facilitate this stage, we combine the data pipelines of Interleave-VLA~\cite{fan2025interleavevlaenhancingrobotmanipulation} and MDT~\cite{reuss2024multimodaldiffusiontransformerlearning} to construct a diverse and hybrid dataset of multimodal goals; (2) \textit{Reflection Tuning}, which unifies scene understanding and action generation, thereby maintaining policy performance while strengthening robust internal reflective reasoning. In the second stage, the model is jointly trained on multimodal data, manipulation tasks, and a corpus constructed following InstructVLA~\cite{yang2025instructvlavisionlanguageactioninstructiontuning}, with additional embodied scenes designed for uncertainty and failure reflection.

Through this design, CollabVLA functions not only as a visuomotor policy in a closed-loop manner, but also as a reflective collaborator that reports reasoning outcomes and proactively incorporates concise textual hints or lightweight visual cues from human input to condition the next action chunk. This avoids reliance on auxiliary world models by \emph{co-dreaming} with humans when facing uncertainty or failures. CollabVLA takes a pinooring step by extending multimodal reasoning in modern VLMs to deeper reflective understanding, while attaining robustness through just-in-time human guidance that steers the success in the long tail. 

In summary, our contributions are as follows:
\begin{itemize}
    \item We identify and systematize the trade-offs among (a) direct autoregressive VLAs, (b) latent-action formulations, and (c) explicit world-model approaches, highlighting the missed opportunity for \emph{lightweight human-in-the-loop} guidance at execution time.
    \item We introduce CollabVLA, a collaborative VLA framework with MoE adaptation that transforms a standard visuomotor policy into a proactive assistant capable of reasoning, acting, and interacting with humans.
    \item We show that CollabVLA improves success rates and maintains low latency, and effectively extends its self-reflection to solicit just-in-time human guidance.

\end{itemize}

\section{Related Work}

\subsection{Vision–Language–Action Models}

The success of VLMs~\cite{alayrac2022flamingovisuallanguagemodel,zhai2023sigmoidlosslanguageimage} has motivated extensions to VLA policies by fine-tuning on robot datasets and casting control as multimodal sequence modeling. Early systems such as RT-1~\cite{brohan2023rt1roboticstransformerrealworld} and OpenVLA~\cite{kim2024openvlaopensourcevisionlanguageactionmodel} adopt autoregressive next-token prediction on robot data, but this erodes multimodal grounding and limits generalization to novel settings or long-horizon tasks. To mitigate this, some work reintroduces large-scale image-text corpora and co-trains them with robot experience to preserve grounding while aligning it with control~\cite{brohan2023rt2visionlanguageactionmodelstransfer}. These approaches further explore latent-action formulations that distill intermediate representations or temporally chunked primitives from unlabeled internet data~\cite{liang2025clamcontinuouslatentaction,chen2025villaxenhancinglatentaction}. Yet most methods remain \emph{black boxes}, which hinders failure analysis and interactive correction.

\begin{figure}[t]
    \centering
    \includegraphics[width=1\linewidth]{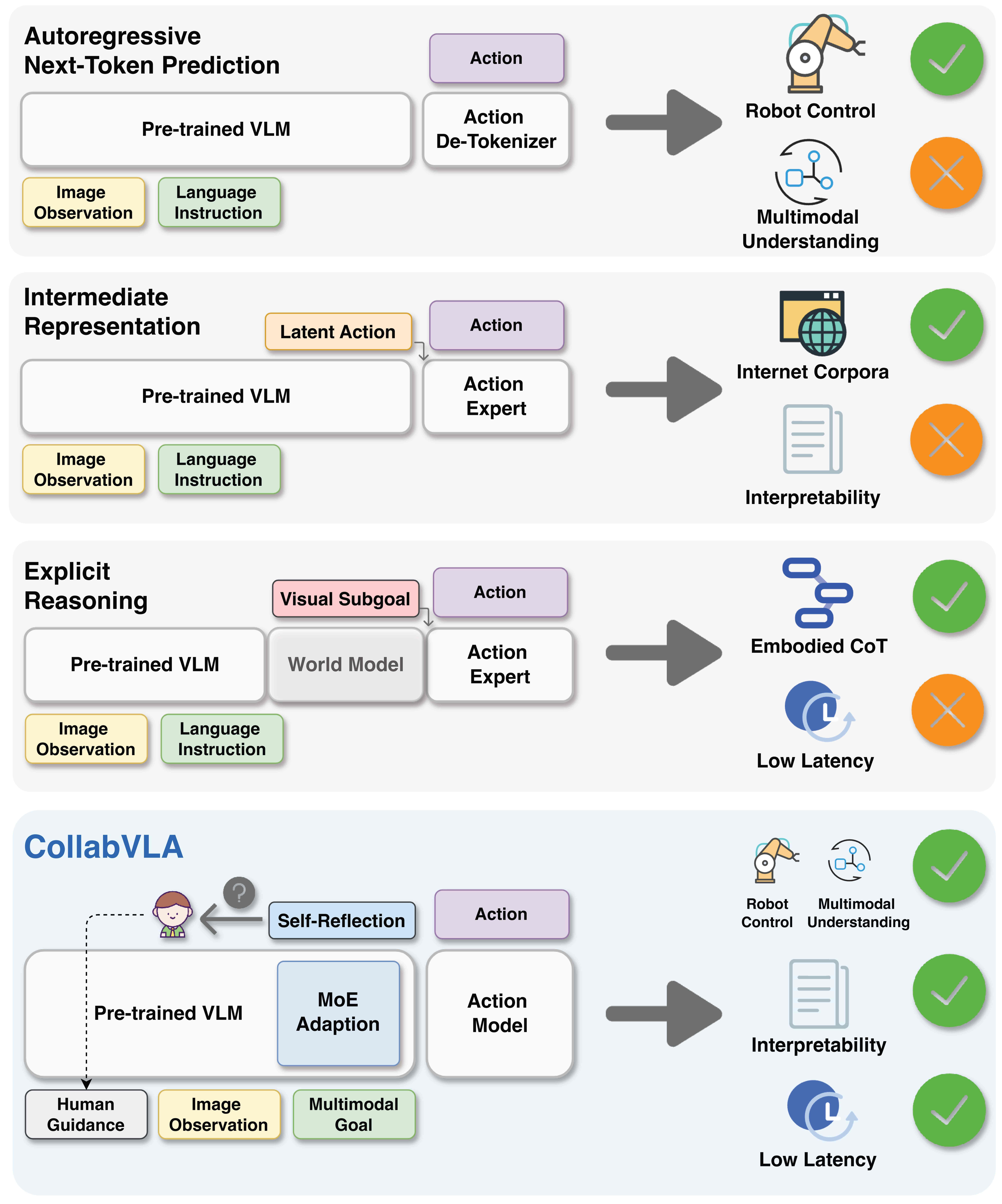}
    \caption{\textbf{Comparison between prior methods and CollabVLA.} Prior methods often lose multimodal grounding, lack interpretability, or fail on unseen predictions and come up with high latency. In contrast, CollabVLA integrates self-reflection with lightweight human guidance to achieve robustness, transparency, and efficiency.}
    \vspace{-6mm}
    \label{fig:related work}
\end{figure}

A subsequent wave of research makes intermediate reasoning explicit. Some methods adopt explicit intermediate or conversational planning, such as ECoT~\cite{zawalski2025roboticcontrolembodiedchainofthought} and ChatVLA~\cite{zhou2025chatvlaunifiedmultimodalunderstanding}. Others rely on visual subgoals, including CoT-VLA~\cite{zhao2025cotvlavisualchainofthoughtreasoning}. Alternatively, diffusion-based or world-model–driven designs, such as GR-MG~\cite{li2024grmgleveragingpartiallyannotated} and RoboDreamer~\cite{zhou2024robodreamerlearningcompositionalworld}, generate imagined future states to guide control. By exposing intermediate plans or auxiliary predictions, they improve transparency and robustness. However, explicit subgoal generation often struggles to generalize to unseen layouts and introduces latency. Most reasoning also remains limited to superficial rationalization, lacking the depth required for insightful guidance. These limitations motivate lightweight human-in-the-loop interaction, which CollabVLA enables by integrating self-reflection with action generation, transforming a visuomotor policy into a collaborative assistant capable of both acting and asking (see Fig.~\ref{fig:related work}).

\subsection{Human-in-the-Loop Collaboration}

Human-in-the-loop (HITL) strategies have been extensively explored for determining when and what to ask humans. Notable approaches addressing uncertainty include KnowNo with conformal prediction~\cite{ren2023robotsaskhelpuncertainty} and Introspective Planning for safety~\cite{liang2025introspectiveplanningaligningrobots}. In parallel, LLM-based critique methods, such as Reflexion~\cite{shinn2023reflexionlanguageagentsverbal}, LLM-as-a-Judge~\cite{zheng2023judgingllmasajudgemtbenchchatbot}, and critic-embedded AssistantX framework~\cite{sun2025assistantxllmpoweredproactiveassistant}, show that self-reflection can be extended to trigger targeted human queries. However, most existing approaches rely on modular pipelines that combine high-level planners with low-level skill libraries, rather than unified models capable of integrated reasoning and control.

Nonetheless, they confirm that pretrained LLMs can act as reflective validators, reasoning over current observations and past progress. Prior embodied studies have partially recognized this by introducing explicit reasoning for self-guidance, but, to our knowledge, CollabVLA is the first VLA to retain a single-backbone visuomotor policy while enabling native reflective reasoning to guide action generation and incorporating real-time human guidance during execution.

\begin{figure*}[t]
    \centering
    \includegraphics[width=1\linewidth]{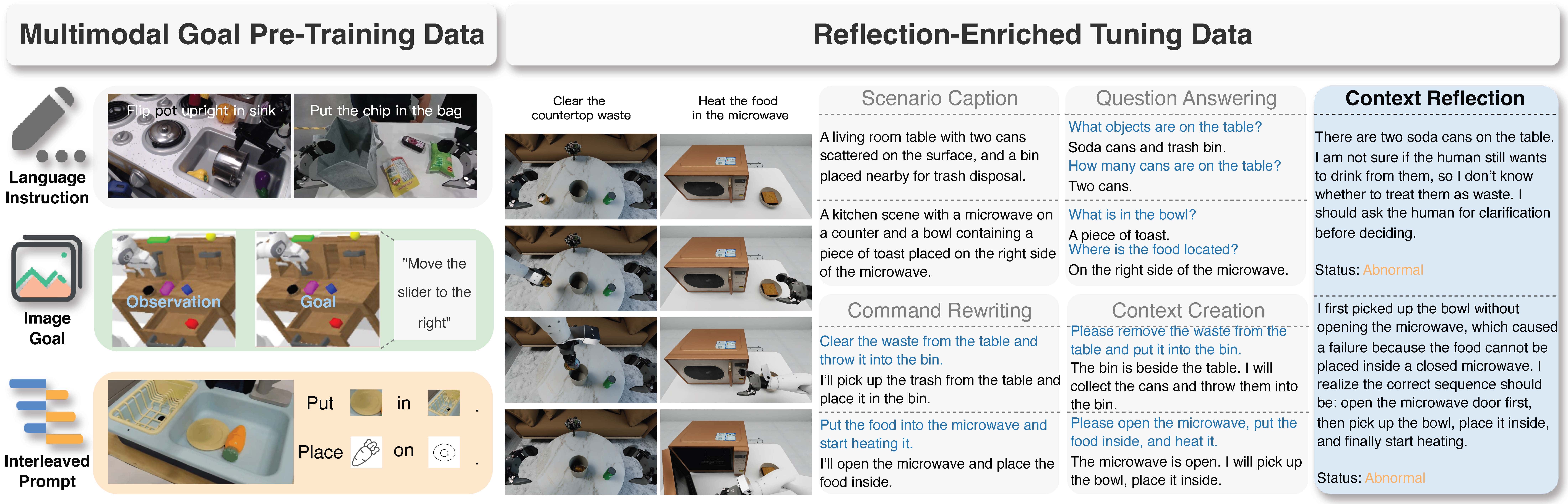}
    \caption{\textbf{Overview of curated datasets for CollabVLA.} The left panel illustrates multimodal goal pre-training data with interleaved prompts and goal image augmentation, whereas the right panel presents reflection-enriched tuning data reformulated as \textit{Context Reflection}, including failures due to inconsistencies between actions and states (e.g., grasping the food before opening the microwave) and ambiguities arising from multiple plausible targets (e.g., two similar cans with unclear intent).}
    \vspace{-3mm}
    \label{fig:data}
\end{figure*}

\section{Methodology}
In this section, we first formalize the problem (Sec.~\ref{method: problem}), then detail the construction of CollabVLA from three components: data (Sec.~\ref{method: data}), model architecture (Sec.~\ref{method: model}), and training pipeline (Sec.~\ref{method: training}).

\subsection{Problem Formulation}
\label{method: problem}

We formalize CollabVLA as a goal-conditioned vision--language--action policy that outputs not only an action sequence but also explicit reflective reasoning and optional human queries. Given a current state $o^t$ (image observations), past states $o^{t-1}$, proprioception $p^t$, and a multimodal goal $g$, the policy $\pi_\theta$ predicts:
\begin{equation}
\pi_\theta(o^t, o^{t-1}, p^t, g) \rightarrow \{\bar{a}^t, r^t, q^t\}
\end{equation}
where $\bar{a}^t = (a^t_i, \dots, a^t_{i+k-1})$ denotes a short action chunk, $r^t$ is a reflective reasoning trace, and $q^t$ is a binary query indicator. The query token $q^t \in \{0,1\}$ specifies whether to solicit human input: $q^t = 0$ allows the agent to proceed autonomously, while $q^t = 1$ triggers a follow-up question.

\subsection{Data}
\label{method: data}

VLA training data are largely language-only or only weakly aligned across vision, language, and action, with sparse, non-instructional labels; these limitations impede free-form instruction following and multimodal goal grounding. Prior work also lacks explicit supervision for \emph{when} to self-diagnose, \emph{when} to ask, and \emph{how} to revise. We therefore curate two complementary corpora as illustrated in Fig.~\ref{fig:data}.

\vspace{1mm}
\noindent\textbf{Multimodal Goal Pre-training.}
Built on simulation and real-world manipulation datasets~\cite{walke2024bridgedatav2datasetrobot, embodimentcollaboration2025openxembodimentroboticlearning, khazatsky2025droidlargescaleinthewildrobot, agibotworldcontributors2025agibotworldcolosseolargescale, liu2023liberobenchmarkingknowledgetransfer, li2024evaluatingrealworldrobotmanipulation}, we add two augmentations: 
(i) \emph{interleaved multimodal prompts} (as in Interleave-VLA~\cite{fan2025interleavevlaenhancingrobotmanipulation}) that reformulate demonstrations into mixed text--image instructions; and 
(ii) \emph{goal-image augmentation} (inspired by MDT~\cite{reuss2024multimodaldiffusiontransformerlearning} and GR-MG~\cite{li2024grmgleveragingpartiallyannotated}) that samples future frames as explicit visual goals. 
We further adopt Diffusion-VLA-style reasoning augmentation~\cite{wen2025diffusionvlageneralizableinterpretablerobot} to attach concise language rationales to trajectories, injecting planning-oriented signals that act as lightweight self-reflection for successful cases.

\vspace{1mm}
\noindent\textbf{Reflection-Enriched Tuning.}
To couple action with reflective behavior, we extend the InstructVLA pipeline~\cite{yang2025instructvlavisionlanguageactioninstructiontuning} with a \emph{Context Reflection} task: the agent explains past/current observations and diagnoses uncertainty or failure. 
We synthesize hard cases by (i) inserting irrelevant or shuffled frames to break temporal consistency, 
(ii) adding key objects to induce perceptual ambiguity and multiple choices, and 
(iii) perturbing action labels or goal descriptions to simulate failure trajectories. 
Each synthetic sample is formatted as a \texttt{\{observation sequence, instruction, action trace, reflection\}} tuple, 
where the reflection field captures how the agent should articulate uncertainty.
We follow a reflection-oriented data generation pipeline: environment rollouts are paired with large language models to generate natural reflection answers, while visual states are synthesized with generative models and subsequently verified by human annotators for correctness. In addition, we interleave synthetic and real failure cases to balance distribution, and vary reflection styles to add diversity.

\begin{figure*}[t]
    \centering
    \includegraphics[width=1\linewidth]{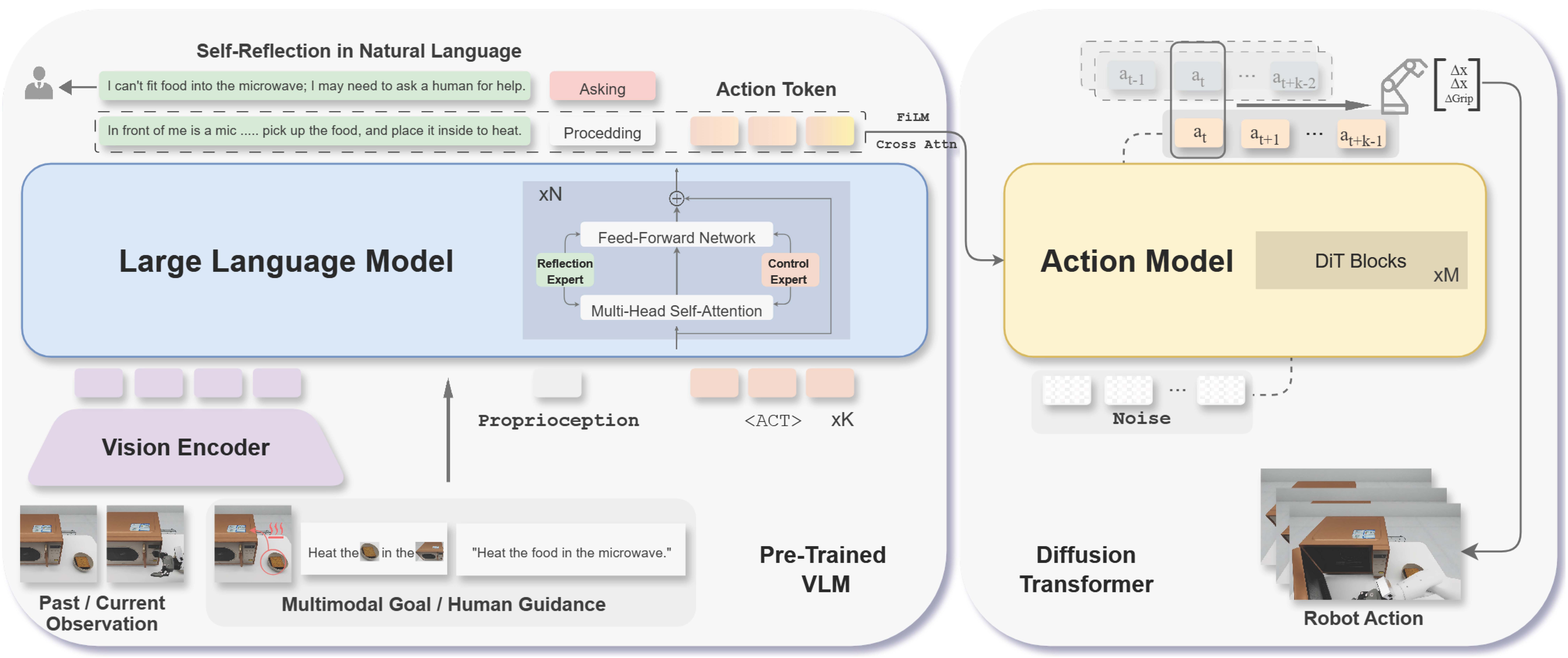}
  \caption{\textbf{Overall architecture of CollabVLA.} The VLM backbone is augmented with LoRA-based \emph{Control} and \emph{Reflection Experts}, adaptively gated inside each Transformer block. Reflection provides natural-language reasoning and human queries, while latent action tokens condition a DiT via cross-attention by being injected as key–value memories that the denoiser queries at each step to retrieve fine-grained intent. The reflection embedding modulates all DiT layers through FiLM.}

   \vspace{-3mm}
    \label{fig:method}
\end{figure*}

\subsection{Model Architecture}
\label{method: model}

CollabVLA couples a VLM backbone with MoE adaptation and a diffusion-based action expert for conditional, low-latency trajectory generation (see Fig.~\ref{fig:method}).

\vspace{1mm}
\noindent\textbf{VLM Backbone.}
We build on InternVL2.5~\cite{chen2025expandingperformanceboundariesopensource}, which natively supports image--text interleaving. The backbone consumes a single sequence that concatenates: (i) robot observations (current RGB $o^t$ and optionally a past frame $o^{t-1}$) tagged as ``\texttt{[NOW]}'' and ``\texttt{[PAST]}''; cached embeddings for $o^{t-1}$ are reused to reduce latency; (ii) a multimodal goal $g$ plus optional human guidance wrapped by ``\texttt{<HumanTip>} ... \texttt{</HumanTip>}''; (iii) proprioception $p^t$ projected by a small MLP; and (iv) $K$ learnable \texttt{[ACT]} queries. The model outputs: (a) a reflection string; (b) a binary ask indicator from a classifier over the pooled reflection representation; and (c) $K$ latent action embeddings from the final hidden states of the \texttt{[ACT]} queries, which seed the action expert.

\vspace{1mm}

\noindent\textbf{MoE Adaption.} Inside the backbone, we introduce a \emph{Mixture-of-Experts design} that enables the model to alternate between reflection and forward planning. Unlike ChatVLA~\cite{zhou2025chatvlaunifiedmultimodalunderstanding}, which uses static routing to activate either a control or a conversational FFN, we adopt an adaptive gating scheme similar to InstructVLA~\cite{yang2025instructvlavisionlanguageactioninstructiontuning}. We insert LoRA experts~\cite{hu2021loralowrankadaptationlarge} into the linear projections of MHA and FFN: a \emph{Control Expert} and a \emph{Reflection Expert}. For hidden state $x$, the output is: \begin{equation} h = W_{0}x + (B_{\text{ctrl}}A_{\text{ctrl}}x)\alpha_{\text{ctrl}}\lambda_{\text{ctrl}} + (B_{\text{ref}}A_{\text{ref}}x)\alpha_{\text{ref}}\lambda_{\text{ref}} \end{equation} where $W_{0}$ is the frozen pretrained weight, $A_{\cdot}\in\mathbb{R}^{r\times d}$ and $B_{\cdot}\in\mathbb{R}^{d\times r}$ are LoRA parameters, $\alpha_{\cdot}$ are scaling factors, and $\lambda_{\cdot}$ are gating coefficients predicted by: \begin{equation} \lambda = \text{softmax}\!\Big(W_{g}\sum_j \text{softmax}(w^\top h_j)\,h_j\Big) \end{equation} where $W_{g}$ is a lightweight gating head, $h_j$ are token hidden states of the current layer, and $w$ is a learned attention vector. This adaptive gating favors the Control Expert during routine control and shifts to the Reflection Expert under uncertainty, allowing the model to balance reasoning and acting.

\vspace{1mm}

\noindent\textbf{Diffusion-Based Action Model.}
We employ a Diffusion Transformer (DiT)~\cite{peebles2023scalablediffusionmodelstransformers} as the action generator. The VLM backbone provides two signals: (i) \emph{latent action tokens} from learnable \texttt{[ACT]} queries, encoding structured intentions and injected as key--value memories in cross-attention for fine-grained trajectory control; and (ii) a \emph{reflection embedding}, the final hidden states of reflection tokens, broadcast via FiLM~\cite{perez2017filmvisualreasoninggeneral} to modulate hidden activations with global semantic guidance. Starting from Gaussian noise in action space, the DiT iteratively refines trajectories conditioned on both signals, yielding outputs that are dynamically consistent, physically feasible, and reasoning-aligned.

\vspace{1mm}

\noindent\textbf{Inference.}
During inference, CollabVLA executes a concise \emph{reflect–ask/act} two-pass loop. The backbone first decodes a short reflection, pools it into an embedding, and a binary head predicts the ask indicator $\hat{q}\!\in\!\{0,1\}$. If $\hat{q}\!=\!1$, the reflection (which already includes the uncertainty information) is shown to the user; the reply is appended to the goal and a second forward pass is run. If $\hat{q}\!=\!0$, the reflection embedding and $K$ latent \texttt{[ACT]} queries are sent directly to the diffusion expert. To reduce latency, we support (i) stopping autoregressive decoding once the first \texttt{[ACT]} appears and decoding the remaining queries in parallel, and (ii) caching the reflection and \texttt{[ACT]} tokens across steps. To smooth control when new guidance arrives, we blend the current action with cached predictions using a similarity-weighted average~\cite{li2024cogactfoundationalvisionlanguageactionmodel}:
\begin{equation}
\hat{a}_t \;=\; \frac{\sum_{k=0}^{K} \exp\!\big(\alpha\,\langle a_t(o_t),\, a_t(o_{t-k})\rangle\big)\, a_t(o_{t-k})}
{\sum_{j=0}^{K} \exp\!\big(\alpha\,\langle a_t(o_t),\, a_t(o_{t-j})\rangle\big)}
\end{equation}
where $\langle\cdot,\cdot\rangle$ is cosine similarity and $\alpha\!=\!0.1$. This yields smooth, mode-consistent trajectories with negligible overhead.

\subsection{Training Pipeline}
\label{method: training}

We adopt a two-stage recipe that first grounds visuomotor control and then enhances reflective reasoning, ensuring that reflection does not compromise action performance.

\vspace{1mm}
\noindent\textbf{Action Grounding.} This stage aligns multimodal perception with motor actions and lightweight planning language, enabling the agent to describe and execute task steps. Training uses the multimodal goal pre-training data in Sec.~\ref{method: data}, which provide paired observations, multimodal goals, trajectories, and rationales, while only the Control Expert is activated. The VLM backbone with Control-LoRA $\theta_{\text{ctrl}}$ outputs a planning string $\hat{r}_t$ and a set of latent action tokens $z$ from learnable \texttt{[ACT]} queries. Unlike UniVLA~\cite{bu2025univlalearningacttaskcentric}, which first trains a separate latent-action model to annotate demonstrations, here $z$ has no explicit ground-truth labels: it is learned implicitly as a conditioning signal for the diffusion action model to reconstruct ground-truth trajectories. The overall loss is:
\begin{equation}
\mathcal{L}_{\text{Stage1}} = \lambda_{\text{lang}} \cdot \mathcal{L}_{\text{lang}} + \lambda_{\text{act}} \cdot \mathcal{L}_{\text{diff}}
\end{equation}
where:
\begin{equation}
\mathcal{L}_{\text{lang}} = -\!\!\!\sum_{j} \log P_{\theta_{\text{ctrl}}}(r_{t,j}\,|\,o^t,o^{t-1},p^t,g)
\end{equation}
is the cross-entropy loss on forward planning traces, and: 
\begin{equation} \mathcal{L}_{\text{diff}} = \mathbb{E}_{s,\epsilon}\Big[ \big\| \epsilon - \varepsilon_\phi(\mathbf{x}_s, z, s) \big\|^2 \Big], \quad \mathbf{x}_s = \alpha_s \tau_t + \sigma_s \epsilon \end{equation} 
is the diffusion denoising loss on robot trajectories (with $\tau_t$ the ground-truth trajectory at time $t$). Here, $\mathcal{L}_{\text{lang}}$ updates the Control-Expert LoRA $\theta_{\text{ctrl}}$, while $\mathcal{L}_{\text{diff}}$ updates the DiT parameters $\phi$ and $\theta_{\text{ctrl}}$.

\vspace{1mm}
\noindent\textbf{Reflection Tuning.} In this stage, we freeze the diffusion action model and the VLM backbone, while jointly training the two VLM-side LoRA experts and the auxiliary heads. Training leverages a hybrid corpus that combines diverse multimodal tasks (e.g., VQA, captioning, referring expressions, retrieval, commonsense reasoning, and manipulation tasks) with the reflection-enriched data described in Sec.~\ref{method: data}. The objective is to unify action generation with reflective reasoning, particularly when the policy encounters failures or ambiguities, enabling the model to not only execute but also assess, revise, and query. Let $\theta_{\text{ctrl}}$ and $\theta_{\text{refl}}$ denote the Control- and Reflection-Expert LoRAs on the VLM, $\psi$ the ask-indicator head, and $W_{\text{gate}}$ the lightweight gating network. The overall training loss is:
\begin{equation}
\mathcal{L}_{\text{Stage2}} = \lambda_{\text{ref}} \cdot \mathcal{L}_{\text{ref}} + \lambda_{\text{ask}} \cdot \mathcal{L}_{\text{ask}} 
\end{equation}
Here, $\mathcal{L}_{\text{ref}}$ is the standard token-level cross-entropy for decoding the reflection string across the above task mixture, and:
\begin{equation}
\mathcal{L}_{\text{ask}} = -\Big[ y \log \hat{y} + (1-y)\log(1-\hat{y}) \Big], \quad y \in \{0,1\}
\end{equation}
is the binary cross-entropy for the ask-indicator head. $\mathcal{L}_{\text{ref}}$ updates $(\theta_{\text{ctrl}}, \theta_{\text{refl}}, W_{\text{gate}})$; $\mathcal{L}_{\text{ask}}$ updates \emph{only} the ask head $\psi$. The VLM backbone and DiT parameters remain frozen. This enables explicit reflective reasoning, calibrated uncertainty recognition, and human-query triggering, without degrading the visuomotor competence established in the former stage.

\input{table/multimodal}

\input{table/simulation}

\section{Experiments}
We empirically evaluate CollabVLA through four questions: \emph{(1)} Does it preserve strong multimodal understanding? \emph{(2)} Can it detect ambiguity and execution failures and produce insightful reflections? \emph{(3)} How well can it interpret human-in-the-loop feedback to improve performance on long-horizon, complex tasks? \emph{(4)} To what extent can it deliver these gains with minimal human effort and latency overhead? We benchmark across simulation and real-world tasks, and compare against state-of-the-art VLA baselines and ablations.

\vspace{1mm}

\noindent \textbf{Summary of findings.} CollabVLA preserves—and often exceeds—multimodal understanding relative to strong VLA baselines (Q1). During execution, it identifies ambiguities and failures, producing concise reflections that condition the action model or trigger human queries (Q2). With brief, free-form human hints, it improves long-horizon success while keeping latency low (Q3). Compared with strong baselines—including those that rely on heavy generative detours—it delivers a better effectiveness–efficiency trade-off (Q4). These results underscore CollabVLA as a practical, collaborative assistant that unifies perception, action, and collaboration.

\subsection{Experimental Setup}

\noindent \textbf{Environments and Tasks.}
We evaluate along three axes. 
(i) \emph{Comprehensive understanding}: we report on four multimodal-understanding benchmarks (MMMU, MMStar, OCRBench, HallBench) and four VQA benchmarks (TextVQA, DocVQA, InfoVQA, RealWorldQA), plus a 500-example \emph{ContextReflection} set constructed from AgibotWorld~\cite{agibotworldcontributors2025agibotworldcolosseolargescale}. The ContextReflection set spans real-world executions and GenieSim scenes and is held out from reflection-tuning data. 
(ii) \emph{Simulation}: we extend the Simpler setting~\cite{li2024evaluatingrealworldrobotmanipulation} into \textit{Simpler-Collab}, a 200-task suite covering 8 task types (see Table~\ref{tab:simpler_table2}) focused on long-horizon control and ambiguity resolution. We lengthen horizons, add controlled ambiguity and multiple-choice subgoals, and provide each task with a hidden script specifying environment, objects, and goals. When the VLA queries for help, its question and the script are passed to an LLM simulating the human, enabling automated human-in-the-loop evaluation.
(iii) \emph{Real-world tasks}: we use a DOBOT CR5 arm equipped with a ROBOTIQ gripper and a UR5 with an AG95 gripper. The benchmark spans five task categories with four instances each; details are provided in Sec.~\ref{real}.

\vspace{1mm}
\noindent \textbf{Comparisons.} We compare to SOTA VLAs and ablations:

\begin{itemize}
    \item \textit{Vanilla VLA:} OpenVLA~\cite{kim2024openvlaopensourcevisionlanguageactionmodel} generates action tokens autoregressively. We also build OpenVLA-Collab by fine-tuning OpenVLA on our robotic and multimodal data (following the authors’ recipe) and adding only a binary ask head; reflections are decoded with the native LM head, leaving the perception–action stack unchanged.
    \item \textit{Hierarchical VLAs:} $\pi$0~\cite{black2024pi0visionlanguageactionflowmodel} and UniVLA~\cite{bu2025univlalearningacttaskcentric}, which explicitly decouple perception and control. We fine-tune them on our training data and add an ask head to obtain Collab variants; reflections are prompted from the LM head, and the action modules are unchanged.
    \item \textit{Explicit reasoning VLAs:} ECoT~\cite{zawalski2025roboticcontrolembodiedchainofthought}, CoT-VLA~\cite{zhao2025cotvlavisualchainofthoughtreasoning}, ChatVLA~\cite{zhou2025chatvlaunifiedmultimodalunderstanding}, DiVLA~\cite{wen2025diffusionvlageneralizableinterpretablerobot}, InstrcutVLA~\cite{yang2025instructvlavisionlanguageactioninstructiontuning}, and RoboDreamer~\cite{zhou2024robodreamerlearningcompositionalworld}. We do not build Collab variants; instead, we assess their self-generated rationales and latency under identical scenes, contrasting with CollabVLA’s strategy of querying humans only when appropriate.
    \item \textit{Ablations:}
    (i) \textbf{No-Tuning} (only Stage~1 data is used);
    (ii) \textbf{No-Ref} (no context reflection data);
    (iii) \textbf{No-FiLM} (reflections are generated but not used to condition the action model);
    (iv) \textbf{No-Ask} (reflection conditions the action model but question triggering is disabled);
    (v) \textbf{No-MoE} (the dual-expert LoRA design is removed, and a single VLM LoRA is trained and shared across Stage~1 and Stage~2);
    (vi) \textbf{No-MG} (no multimodal goals).
    
\end{itemize}

\begin{figure*}[t]
    \centering
    \includegraphics[width=1\linewidth]{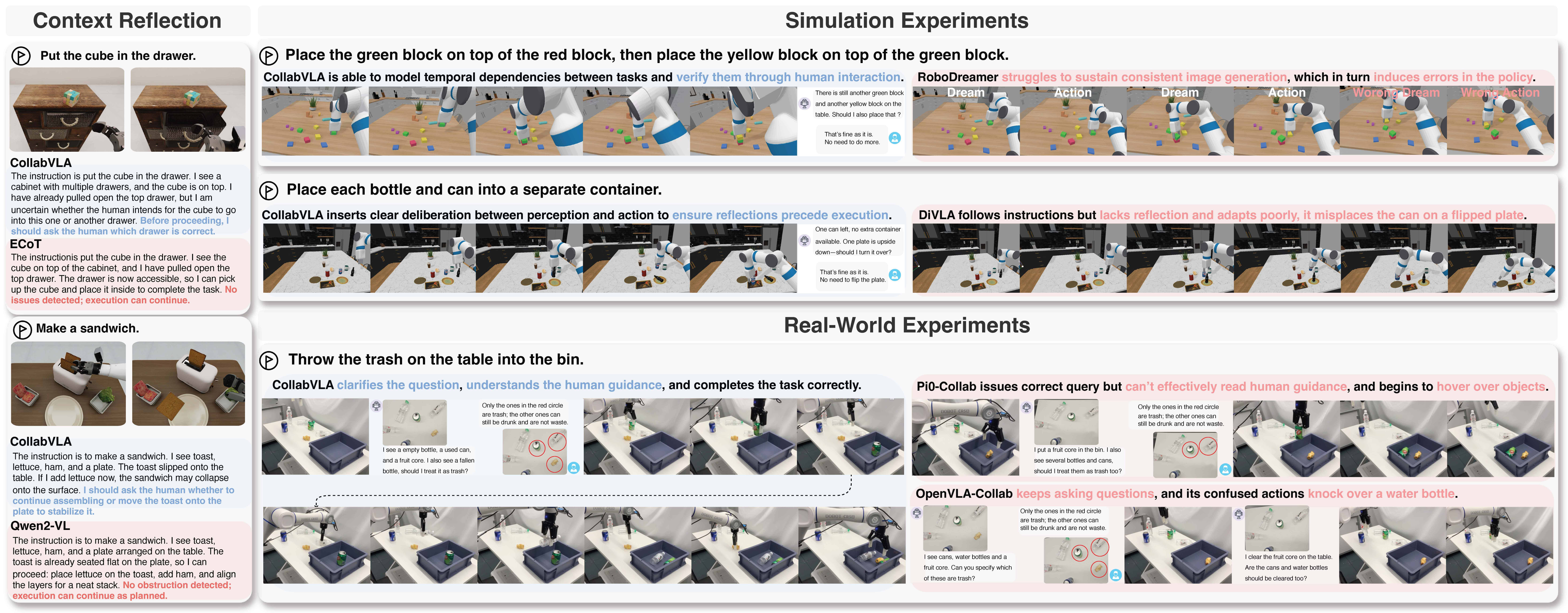}
 \caption{\textbf{Demonstration of experiment results.} The left context-understanding task shows that CollabVLA not only handles multiple-choice reasoning better than ECoT~\cite{zawalski2025roboticcontrolembodiedchainofthought}, but also detects action–observation gaps to prevent long-run execution errors, outperforming general MLLMs like Qwen2-VL~\cite{wang2024qwen2vlenhancingvisionlanguagemodels} in embodied settings. On the right, in complex, unseen stacking setting, RoboDreamer~\cite{zhou2024robodreamerlearningcompositionalworld} struggles because its generation generalize poorly and its actions rely on inverse dynamics from generation. CollabVLA leverages self-reflection and timely oracle human guidance to refine its policy. The real-robot results further illustrate CollabVLA’s superior ability to interpret multimodal human guidance and translate it into effective strategies compared with Collab-variants of $\pi$0~\cite{black2024pi0visionlanguageactionflowmodel} (which asks but struggles to translate well) and OpenVLA ~\cite{kim2024openvlaopensourcevisionlanguageactionmodel} (which over-asks).}

   \vspace{-5mm}
    \label{fig:experiment}
\end{figure*}

\subsection{Main Results}

\noindent \textbf{Multimodal understanding and contextual reflection.}
Across comprehensive understanding and VQA task, \emph{CollabVLA} matches/exceeds strong VLAs and remains competitive with larger MLLMs (see Table~\ref{tab:multimodal_eval}).
Concretely, CollabVLA attains \textbf{49.0}/\textbf{57.2} on MMMU$^{\text{Val}}$/MMStar\textendash near InternVL2.5 (51.8/58.7) and LLaVA-OV (47.9/61.9)\textendash scores \textbf{814} on OCRBench (vs.\ 820 best) and \textbf{43.7} on HallBench (vs.\ 46.6 best), and achieves \underline{\textbf{88.6}} on \textsc{ConRef}, best overall.
Against the strongest VLA baseline (InstructVLA): +4.8/+1.6/$-2$/+0.3/+26.6 on the above metrics.
On VQA, CollabVLA leads \emph{two of four} benchmarks\textendash TextVQA \underline{\textbf{77.0}} and InfoVQA \underline{\textbf{65.1}}\textendash and remains competitive on DocVQA (85.7 vs.\ 88.6, Qwen2-VL) and RWQA (65.3 vs.\ 69.9, LLaVA-OV).
Thus, our two-stage training \emph{does not sacrifice} multimodal competence and competes with strong VLMs; whereas training a VLM \emph{only} for actions can catastrophically forget general skills (OpenVLA reports \textbf{0.0} across understanding/VQA).
We attribute CollabVLA’s gains to Stage-2 training, which teaches the model to detect uncertainty and compose evidence (e.g., scene perception, spatial/temporal resolution) with \emph{robust, calibration-aware} reflections, yielding sharper grounding that transfers to diverse multimodal tasks.

\noindent
\textbf{Control and efficiency.}
On \textit{Simpler\textendash Collab}, \emph{CollabVLA} attains the best SR in \emph{all 8} subtasks (Table~\ref{tab:simpler_table2}). It also executes compactly (\textit{Time}/\textit{Dream}=\textbf{36}/\textbf{\underline{1.9}}), while collab baselines help but remain less efficient: OpenVLA\textendash Collab 90/7.2, $\pi$0\textendash Collab 44/3.6, UniVLA\textendash Collab 49/4.4. Pure explicit\textendash reasoning agents are slower and intervene more (RoboDreamer 94/17.2, GR\textendash MC 74/9.8), showing that \emph{selective} human queries are more cost\textendash effective than long rationales. \emph{CollabVLA} instead asks sparingly (1.9 queries/episode; $\sim$2--4$\times$ fewer than other collab variants) and runs faster: \textit{OpenVLA} is slow (81/\textemdash) due to token\textendash by\textendash token action decoding; \textit{RoboDreamer} (94/17.2) pays for image generation and inverse dynamics. Overall, \emph{CollabVLA} yields the best success with low \textit{Time} and minimal \textit{Dream}, achieving a superior effectiveness\textendash efficiency trade\textendash off. We present several case studies in Fig.~\ref{fig:experiment} that demonstrate how CollabVLA outperforms other methods.

\input{table/real}

\subsection{Ablation Studies}

\noindent
\textbf{Where does performance drop?}
The largest drop occurs for \textbf{No\textendash Tuning} due to domain overfitting on robot data (e.g., Move: $23.8$ vs. \textbf{\underline{62.2}} for CollabVLA). Removing the dual\textendash expert routing (\textbf{No\textendash MoE}) induces task interference between perception and control and yields similarly large drops (e.g., Move $28.5$ vs. \textbf{\underline{62.2}}). Removing Stage\;2 reflection supervision (\textbf{No\textendash Ref}) or bypassing reflection conditioning (\textbf{No\textendash FiLM}) degrades both SR and rollout quality (e.g., No\textendash FiLM Open/Close $28.9$ vs. \textbf{\underline{63.8}}; Stack $19.1$ vs. \textbf{\underline{42.5}}), confirming that concise, on\textendash policy \emph{reflections} that condition the action are key. Text\textendash only goals (\textbf{No\textendash MG}) remain strong and sometimes rival hierarchical baselines (Pick $55.5$, Move $59.2$) but full multimodal goal grounding still provides a consistent lift. Finally, \textbf{No\textendash Ask} achieves the fastest time (\textit{Time} $=$ \textbf{32}) but underperforms CollabVLA on SR (e.g., Pick $50.8$ vs. \textbf{\underline{58.5}}, Open/Close $55.5$ vs. \textbf{\underline{63.8}}), showing that judicious, sparse human queries are worth the small overhead.

\subsection{Real-World Experiments}
\label{real}
We evaluate five task families on a DOBOT CR5 arm and a UR5 to assess robustness and cross-arm generalization: 
(i) object pick\&place; (ii) open drawer \& store items; (iii) open drawer \& retrieve items; (iv) sort tabletop items; and (v) clear countertop waste. 
Each family has four difficulty tiers with stepwise credit—\emph{Basic} (10), \emph{Distractors} (20), \emph{Clarification} (30), and \emph{Long\textendash horizon} (40). 
We report \emph{SR} as the number of fully completed instances per family ($0\text{--}4$) and \emph{Score} as the summed credit normalized to $[0,100]$. 


From Table~\ref{tab:real}, \emph{CollabVLA} achieves the best average SR/Score (\textbf{\underline{2.5}}/\textbf{\underline{62.2}}), leading in \textbf{4}/5 tasks.
Compared with $\pi$0\textendash Collab, CollabVLA improves the mean by \textbf{+0.2} SR and \textbf{+7.5} Score, with the largest gains on Tasks 2–4 (\(+0.4/+11.3\), \(+0.4/+12.5\), \(+0.3/+2.3\)). 
Against OpenVLA\textendash Collab, the margins widen to \textbf{+1.6} SR and \textbf{+32.4}. 
These trends indicate that CollabVLA’s reflect–ask/act loop with FiLM\textendash conditioned control turns brief human hints into more reliable progress under clutter, ambiguity, and multi\textendash step objectives, while remaining competitive on straightforward manipulation.

\section{Conclusion}
We introduce CollabVLA, a self-reflective VLA reasons explicitly, reflects on uncertainty, and integrates lightweight human feedback in real time. Our two-stage recipe, which first grounds perception in action and then tunes reflection without harming control, yields consistent gains in success and interpretability across tasks. Looking ahead, several high-impact avenues could further amplify CollabVLA’s capabilities, such as integrating tactile and force sensing, advancing epistemic uncertainty with a better-calibrated ask trigger, and evolving collaboration from queries to proactive task allocation and coordination. 

Nonetheless, we are confident that the present instantiation already constitutes a distinctive, practically impactful advance: uniting reflection with action, it delivers consistent gains and supports real-time human collaboration—moving embodied agents toward the next frontier: \textit{robots that not only act, but reflect, adapt, and partner with humans as genuine teammates.}

\bibliographystyle{IEEEtran}
\bibliography{reference}

\end{document}

%% file: table/multimodal.tex
\begin{table*}[t]
\centering
\small
\setlength{\tabcolsep}{5pt}
\caption{\textbf{Results on multimodal-understanding benchmarks, the \emph{ContextReflection} set, and VQA.} We additionally compare against similar-sized MLLMs to quantify how much multimodal ability a VLA retains after learning robotic actions. MLLM numbers are mainly from official reports; VLA numbers from InstructVLA~\cite{yang2025instructvlavisionlanguageactioninstructiontuning} and ChatVLA~\cite{zhou2025chatvlaunifiedmultimodalunderstanding}.}
\label{tab:multimodal_eval}
\begin{tabular}{l c c c c c c c c c c}
\toprule
\multirow{2}{*}{Methods} & \multirow{2}{*}{\#Params} &
\multicolumn{5}{c}{\textbf{Comprehensive Understanding}} &
\multicolumn{4}{c}{\textbf{VQA}} \\
\cmidrule(lr){3-7} \cmidrule(lr){8-11}
 &  & MMMU$^{\text{Val}}$ & MMStar & OCRB & HallB & \textsc{ConRef} & TextVQA & DocVQA & InfoVQA & RWQA \\
\midrule
\multicolumn{11}{c}{\textbf{Multimodal Large Language Models}} \\
\midrule
Qwen2-VL~\cite{wang2024qwen2vlenhancingvisionlanguagemodels} & 2B & 41.1 & 48.0 & 809 & 41.7 & 53.2 & \textbf{74.9} & \textbf{\underline{88.6}} & \textbf{61.4} & 62.9 \\ InternVL2.5~\cite{chen2025expandingperformanceboundariesopensource} & 4B & \textbf{\underline{51.8}} & 58.7 & \textbf{\underline{820}} & \textbf{\underline{46.6}} & 61.4 & $-$ & $-$ & $-$ & $-$ \\ LLaVA-OV~\cite{li2024llavaonevisioneasyvisualtask} & 8B & 47.9 & \textbf{\underline{61.9}} & 622 & 31.6 & \textbf{69.4} & $-$ & $-$ & $-$ & \textbf{\underline{69.9}} \\
Magma~\cite{zawalski2025roboticcontrolembodiedchainofthought}         & 8B & 38.8 & 41.3 & 518  & 38.0 & 64.8 & 66.5 & 65.4 & 45.2 & 56.5 \\
\midrule
\multicolumn{11}{c}{\textbf{Vision–Language–Action Models}} \\
\midrule
DiVLA~\cite{wen2025diffusionvlageneralizableinterpretablerobot}        & 2B & 17.2 & 21.1 & 294  & 9.0  & 39.2 & 7.5  & 15.2 & 14.7 & 25.2 \\
ChatVLA~\cite{zhou2025chatvlaunifiedmultimodalunderstanding}           & 2B & 37.4 & 47.2 & 729  & 39.9 & 54.6 & 2.5  & 29.2 & 43.4 & 47.2 \\
InstructVLA~\cite{yang2025instructvlavisionlanguageactioninstructiontuning} & 2B & \textbf{44.2} & \textbf{55.6} & \textbf{816} & \textbf{43.4} & \textbf{62.0} & \textbf{76.6} & \textbf{85.5} & \textbf{64.7} & \textbf{63.7} \\
OpenVLA~\cite{kim2024openvlaopensourcevisionlanguageactionmodel}      & 7B & 0.0  & 0.0  & 0.0  & 0.0  & 0.0  & 0.0  & 0.0  & 0.0  & 0.0 \\
ECoT~\cite{zawalski2025roboticcontrolembodiedchainofthought}          & 7B & 5.4  & 0.0  & 12   & 0.9  & 34.8 & 0.0  & 0.0  & 0.0  & 0.0 \\
\midrule
CollabVLA \scriptsize(\textbf{No-Tuning}) & 4B & 17.1 & 11.6 & 118 & 12.3 & 16.0 & 9.6 & 13.1 & 11.2 & 33.0 \\
CollabVLA \scriptsize(\textbf{No-MoE})    & 4B & 20.3 & 24.7 & 335 & 18.1 & 29.2 & 10.2 & 18.4 & 16.5 & 58.8 \\
CollabVLA \scriptsize(\textbf{No-Ref})    & 4B & 38.6 & 48.9 & 790 & 41.0 & 34.4 & 68.2 & 76.6 & 65.0 & 53.6 \\
\textbf{CollabVLA}             & 4B & \textbf{49.0} & \textbf{57.2} & \textbf{814} & \textbf{43.7} & \underline{\textbf{88.6}} & \underline{\textbf{77.0}} & \textbf{85.7} & \underline{\textbf{65.1}} & \textbf{65.3} \\
\bottomrule
\end{tabular}
 \begin{tablenotes}    
        \footnotesize          
        \item[*] *Underlined values (\underline{\textbf{x}}) are the best overall in each column; \textbf{bold} indicates the best within each category. A dash ($-$) indicates not officially reported.  
    \end{tablenotes}   
    \vspace{-2mm}
\end{table*}

%% file: table/simulation.tex
\begin{table*}[t]
\centering
\small
\setlength{\tabcolsep}{4pt}
\caption{\textbf{Results on simulation experiments.} We compare VLAs with and without explicit reasoning guidance, along with their collaborative variants. Our observations suggest that the closer an agent comes to resembling human behavior—\textit{reading, reasoning, and interacting in a human-like manner}—the more reliable, adaptive, and effective its actions tend to be.}

\label{tab:simpler_table2}
 \begin{tabular*}{\textwidth}{@{\extracolsep{\fill}} l *{16}{c} c @{}}
\toprule
\multirow{3}{*}{Methods} &
\multicolumn{8}{c}{\textbf{Fetch Robot}} &
\multicolumn{8}{c}{\textbf{WidowX Robot}} &
\multirow{2}{*}{{\footnotesize\itshape Time/Dream}} \\
\cmidrule(lr){2-9} \cmidrule(lr){10-17}
& \multicolumn{2}{c}{Pick} & \multicolumn{2}{c}{Move} & \multicolumn{2}{c}{Open/Close} & \multicolumn{2}{c}{Stack} & \multicolumn{2}{c}{Put} & \multicolumn{2}{c}{Put} & \multicolumn{2}{c}{Stack} & \multicolumn{2}{c}{Put} & \\
& \multicolumn{2}{c}{Item} & \multicolumn{2}{c}{Near} & \multicolumn{2}{c}{Drawer} & \multicolumn{2}{c}{Item} & \multicolumn{2}{c}{Spoon} & \multicolumn{2}{c}{Carrot} & \multicolumn{2}{c}{Block} & \multicolumn{2}{c}{Eggplant} & \\
\cmidrule(lr){2-3}\cmidrule(lr){4-5}\cmidrule(lr){6-7}\cmidrule(lr){8-9}
\cmidrule(lr){10-11}\cmidrule(lr){12-13}\cmidrule(lr){14-15}\cmidrule(lr){16-17}
& sr & len & sr & len & sr & len & sr & len & sr & len & sr & len & sr & len & sr & len & \\
\midrule
OpenVLA~\cite{kim2024openvlaopensourcevisionlanguageactionmodel} 
& 5.4 & 0.09 & 11.2 & 0.17 & 19.1 & 0.24 & 0.0 & 0.05 & 2.0 & 0.04 & 0.0 & 0.01 & 0.0 & 0.02 & 2.9 & 0.05 & \textbf{81}/$-$ \\

OpenVLA-Collab            
& \textbf{27.1} & \textbf{0.39} & \textbf{30.9} & \textbf{0.37} & \textbf{21.9} & \textbf{0.26} & \textbf{11.8} & \textbf{0.22} & \textbf{10.6} & \textbf{0.13} & \textbf{18.3} & \textbf{0.28} & \textbf{12.5} & \textbf{0.21} & \textbf{21.1} & \textbf{0.29} & 90/\textbf{7.2} \\

\midrule

RoboDreamer~\cite{zhou2024robodreamerlearningcompositionalworld}
& 12.0 & 0.21 & 15.5 & 0.23 & 20.3 & 0.33 & 13.2 & 0.23 & 9.8 & 0.22 & 16.1 & 0.25 & 5.0 & 0.11 & 18.0 & 0.24 & 94/17.2 \\

GR\text{-}MC~\cite{li2024grmgleveragingpartiallyannotated}
& 20.2 & 0.26 & 26.8 & 0.38 & 22.0 & 0.32 & 14.5 & 0.29 & 10.6 & 0.28 & 17.8 & 0.33 & 9.0 & 0.17 & 21.5 & 0.31 & 74/\textbf{9.8} \\

MDT~\cite{reuss2024multimodaldiffusiontransformerlearning}
& 24.1 & 0.36 & 28.0 & 0.46 & 23.0 & 0.29 & 24.5 & 0.33 & 20.7 & 0.28 & \textbf{29.5} & \textbf{0.48} & 15.0 & 0.32 & 31.5 & \textbf{0.44} & 56/$-$ \\

DiVLA~\cite{wen2025diffusionvlageneralizableinterpretablerobot} & \textbf{28.0} & \textbf{0.42} & \textbf{32.0} & \textbf{0.50} & \textbf{26.3} & \textbf{0.45} & \textbf{27.6} & \textbf{0.39} & \textbf{23.3} & \textbf{0.38} & 29.4 & 0.43 & \textbf{21.5} & \textbf{0.37} & \textbf{32.0} & 0.42 & \textbf{46}/$-$ \\

\midrule

$\pi$0~\cite{black2024pi0visionlanguageactionflowmodel} 
& 35.0 & 0.41 & 40.8 & 0.49 & 31.5 & 0.38 & 31.6 & 0.39 & 28.2 & 0.32 & 30.6 & 0.37 & 27.7 & 0.33 & 34.4 & 0.37 & \textbf{\underline{30}}/$-$ \\

UniVLA~\cite{bu2025univlalearningacttaskcentric} 
& 34.5 & 0.39 & 38.2 & 0.44 & 29.3 & 0.34 & 30.1 & 0.33 & 33.0 & 0.40 & 31.0 & 0.38 & 28.1 & 0.32 & 34.9 & 0.38 & 36/$-$ \\

$\pi$0\text{-}Collab 
& 49.4 & 0.56 & \textbf{57.3} & \textbf{0.63} & 45.4 & 0.52 & \textbf{38.2} & \textbf{0.43} & 41.4 & 0.47 & \textbf{42.4} & \textbf{0.47} & 36.3 & 0.43 & \textbf{45.1} & \textbf{0.49} & 44/\textbf{3.6} \\

UniVLA\text{-}Collab 
& \textbf{53.2} & \textbf{0.60} & 53.6 & 0.62 & \textbf{58.1} & \textbf{0.60} & 34.8 & 0.41 & \textbf{45.0} & \textbf{0.58} & 39.1 & 0.46 & \textbf{39.6} & \textbf{0.44} & 41.4 & 0.46 &  49/4.4 \\

\midrule
CollabVLA {\scriptsize (\textbf{No-Tuning})}
& 18.5 & 0.25 & 23.8 & 0.33 & 15.7 & 0.26 & 12.5 & 0.23 & 13.2 & 0.19 & 11.6 & 0.25 & 10.3 & 0.21 & 19.7 & 0.35 & 53/$-$ \\

CollabVLA \scriptsize(\textbf{No-MoE})
& 23.0 & 0.28 & 28.5 & 0.36 & 21.0 & 0.32 & 16.2 & 0.25 & 18.5 & 0.26 & 24.1 & 0.29 & 18.0 & 0.26 & 29.2 & 0.37 & 49/$-$ \\

CollabVLA \scriptsize(\textbf{No-Ref})
& 27.1 & 0.33 & 29.3 & 0.41 & 22.8 & 0.28 & 16.9 & 0.26 & 20.0 & 0.27 & 23.2 & 0.30 & 20.4 & 0.26 & 29.1 & 0.36 & 37/$-$ \\

CollabVLA \scriptsize(\textbf{No-FiLM})
& 34.8 & 0.45 & 39.2 & 0.44 & 28.9 & 0.35 & 20.0 & 0.28 & 14.4 & 0.24 & 24.0 & 0.34 & 19.1 & 0.29 & 26.2 & 0.42 & 34/2.8 \\

CollabVLA \scriptsize(\textbf{No-Ask})
& 50.8 & 0.57 & 51.0 & 0.59 & 55.5 & 0.57 & 32.3 & 0.38 & 32.6 & 0.35 & 36.7 & 0.43 & 27.0 & 0.41 & 38.9 & 0.45 & \textbf{32}/$-$ \\

CollabVLA \scriptsize(\textbf{No-MG})
& 55.5 & 0.63 & 59.2 & 0.66 & 60.8 & 0.63 & 40.5 & 0.46 & 37.1 & 0.41 & 44.6 & 0.50 & 31.5 & 0.48 & 47.2 & 0.52 & 38/2.3 \\

\textbf{CollabVLA}
& \underline{\textbf{58.5}} & \underline{\textbf{0.68}} 
& \underline{\textbf{62.2}} & \underline{\textbf{0.80}} 
& \underline{\textbf{63.8}} & \underline{\textbf{0.76}} 
& \underline{\textbf{43.5}} & \underline{\textbf{0.59}} 
& \underline{\textbf{47.1}} & \underline{\textbf{0.62}} 
& \underline{\textbf{47.5}} & \underline{\textbf{0.63}} 
& \underline{\textbf{42.5}} & \underline{\textbf{0.61}} 
& \underline{\textbf{49.2}} & \underline{\textbf{0.65}} 
& 36/\textbf{\underline{1.9}} \\

\bottomrule
\end{tabular*}
\begin{tablenotes}    
  \footnotesize          
  \item[*] *\textbf{SR} denotes the success rate of tasks, and \textbf{LEN} the average completion length. 
      \textbf{Time/Dream} is computed only on successfully completed tasks. 
    Time is first normalized within each task across all models via clipped percentile scaling and then averaged over tasks:
    $\tilde T^{m,t}=\mathrm{clip}(T^{m,t}_{\mathrm{raw}},\,p^{t}_{5},\,p^{t}_{95})$, 
    $T^{m,t}_{\mathrm{norm}}=\big\lfloor 5+90\cdot\frac{\tilde T^{m,t}-p^{t}_{5}}{p^{t}_{95}-p^{t}_{5}}+0.5\big\rfloor$.
    Dream reports the mean number of explicit reasoning generations for non\textendash Collab models, or human\textendash ask calls for Collab variants.
  For CollabVLAs that did not undergo the full two-stage training and mixture-of-experts, we found self-reflection to be unreliable, so asking is disabled (Dream is left blank). 
  Although \textbf{MDT} and \textbf{DiVLA} are trained with explicit-reasoning supervision (e.g., auxiliary “thought” tokens or future prediction), they do not invoke such traces at inference time; consequently, Dream is not reported for them either. 
  The \textbf{Simpler-Collab} benchmark extends the original \emph{Simpler} setting by emphasizing collaborative visual-matching scenarios with increased task complexity and broader evaluation coverage. Concretely, it introduces longer-horizon executions (50\%), inherent uncertainty (30\%), and richer visual clutter with distracting background items (20\%). It is implemented in ManiSkill3 (on SAPIEN)~\cite{tao2025maniskill3gpuparallelizedrobotics}, with support for the Fetch and WidowX embodiments and the scene assets.
\end{tablenotes}
\vspace{-5mm}
\end{table*}

%% file: table/real.tex
\begin{table}[t]
\centering
\small
\setlength{\tabcolsep}{6pt}
\caption{\textbf{Results on real tasks.} Each method is evaluated for $5$ trials per task on each arm; we report mean of total 10.}
\label{tab:real}
\begin{tabular}{l *{6}{c}}
\toprule
\multirow{2}{*}{Methods} &
\multicolumn{2}{c}{OpenVLA\text{-}Collab} &
\multicolumn{2}{c}{$\pi$0\text{-}Collab} &
\multicolumn{2}{c}{CollabVLA} \\
\cmidrule(lr){2-3}\cmidrule(lr){4-5}\cmidrule(lr){6-7}
& SR & Score & SR & Score & SR & Score \\
\midrule
Task1 & 1.6 & 45.0 & \underline{\textbf{3.4}} & \underline{\textbf{86.0}} & 3.2 & 84.5 \\
Task2 & 0.8 & 34.0 & 2.2 & 55.5 & \underline{\textbf{2.6}} & \underline{\textbf{66.8}} \\
Task3 & 1.1 & 33.7 & 2.4 & 61.7 & \underline{\textbf{2.8}} & \underline{\textbf{74.2}} \\
Task4 & 0.3 & 15.9 & 1.4 & 34.5 & \underline{\textbf{1.7}} & \underline{\textbf{36.8}} \\
Task5 & 0.9 & 20.2 & 1.9 & 36.0 & \underline{\textbf{2.1}} & \underline{\textbf{48.5}} \\
\midrule
\textbf{Avg.} & 0.9 & 29.8 & 2.3 & 54.7 & \textbf{\underline{2.5}} & \textbf{\underline{62.2}} \\
\bottomrule
\end{tabular}
\vspace{-5mm}
\end{table}